\documentclass[twocolumn,10pt]{asme2e}

\usepackage{cite}
\usepackage{amsmath}
\usepackage{amssymb}
\usepackage{algorithmic}
\usepackage{graphicx}
\usepackage{textcomp}
\usepackage{xcolor}
\usepackage{pgfplots}
\usepackage{subfigure}
\usepgfplotslibrary{groupplots}
\usepackage{tikz}
\usepackage{tikzscale}
\usepackage{cleveref}
\usepackage{float}
\usepackage{multirow}
\usepackage{caption}
\usepackage{siunitx}
\usepackage{comment}
\usepackage{booktabs}
\usepackage[square,numbers]{natbib}
\def\BibTeX{{\rm B\kern-.05em{\sc i\kern-.025em b}\kern-.08em
    T\kern-.1667em\lower.7ex\hbox{E}\kern-.125emX}}
    
\crefname{figure}{Fig.}{Fig.}
\Crefname{figure}{Figure}{Figures}
\crefname{equation}{}{}
\Crefname{equation}{Equation}{Equations}

\newcommand{\ie}{\textit{i}.\textit{e}., }
\newcommand{\eg}{\textit{e}.\textit{g}., }
\newcommand{\st}{\text{s.t. }}

\confshortname{ISFA2022}
\conffullname{the 2022 International Symposium on Flexible Automation}

\confdate{July 3-7}
\confyear{2022}
\confcity{Yokohama}
\confcountry{Japan}

\papernum{ISFA2022-031}

\title{Safe Interactive Industrial Robots using\\ Jerk-based Safe Set Algorithm}

\author{Ruixuan Liu, Rui Chen, Changliu Liu \thanks{This work is in part supported by Siemens and Ford Motor Company. The robot arm is donated by FANUC Corporation.} \thanks{Contact author: {\tt ruixuanl, ruic3, cliu6@andrew.cmu.edu}}
    \affiliation{
    Robotics Institute\\
	Carnegie Mellon University\\
	Pittsburgh, PA, USA
    }	
}

\begin{document}

\maketitle    

%%%%%%%%%%%%%%%%%%%%%%%%%%%%%%%%%%%%%%%%%%%%%%%%%%%%%%%%%%%%%%%%%%%%%%
\begin{abstract}
{The need to increase the flexibility of production lines is calling for robots to collaborate with human workers. However, existing interactive industrial robots only guarantee intrinsic safety (reduce collision impact), but not interactive safety (collision avoidance), which greatly limited their flexibility. The issue arises from two limitations in existing control software for industrial robots: 1) lack of support for real-time trajectory modification; 2) lack of intelligent safe control algorithms with guaranteed collision avoidance under robot dynamics constraints. 
To address the first issue, a jerk-bounded position controller (JPC) was developed previously. This paper addresses the second limitation, on top of the JPC. Specifically, we introduce a jerk-based safe set algorithm (JSSA) to ensure collision avoidance while considering the robot dynamics constraints. The JSSA greatly extends the scope of the original safe set algorithm, which has only been applied for second-order systems with unbounded accelerations. The JSSA is implemented on the FANUC LR Mate 200id/7L robot and validated with HRI tasks. 
Experiments show that the JSSA can consistently keep the robot at a safe distance from the human while executing the designated task.
}
\end{abstract}

%%%%%%%%%%%%%%%%%%%%%%%%%%%%%%%%%%%%%%%%%%%%%%%%%%%%%%%%%%%%%%%%%%%%%%
% \begin{nomenclature}
% \entry{A}{You may include nomenclature here.}
% \entry{$\alpha$}{There are two arguments for each entry of the nomemclature environment, the symbol and the definition.}
% \end{nomenclature}

% The spacing between abstract and the text heading is two line spaces.  The primary text heading is  boldface in all capitals, flushed left with the left margin.  The spacing between the  text and the heading is also two line spaces.

%%%%%%%%%%%%%%%%%%%%%%%%%%%%%%%%%%%%%%%%%%%%%%%%%%%%%%%%%%%%%%%%%%%%%%
\section{Introduction}

\begin{figure}
\footnotesize
\subfigure[]{\includegraphics[width=0.48\linewidth]{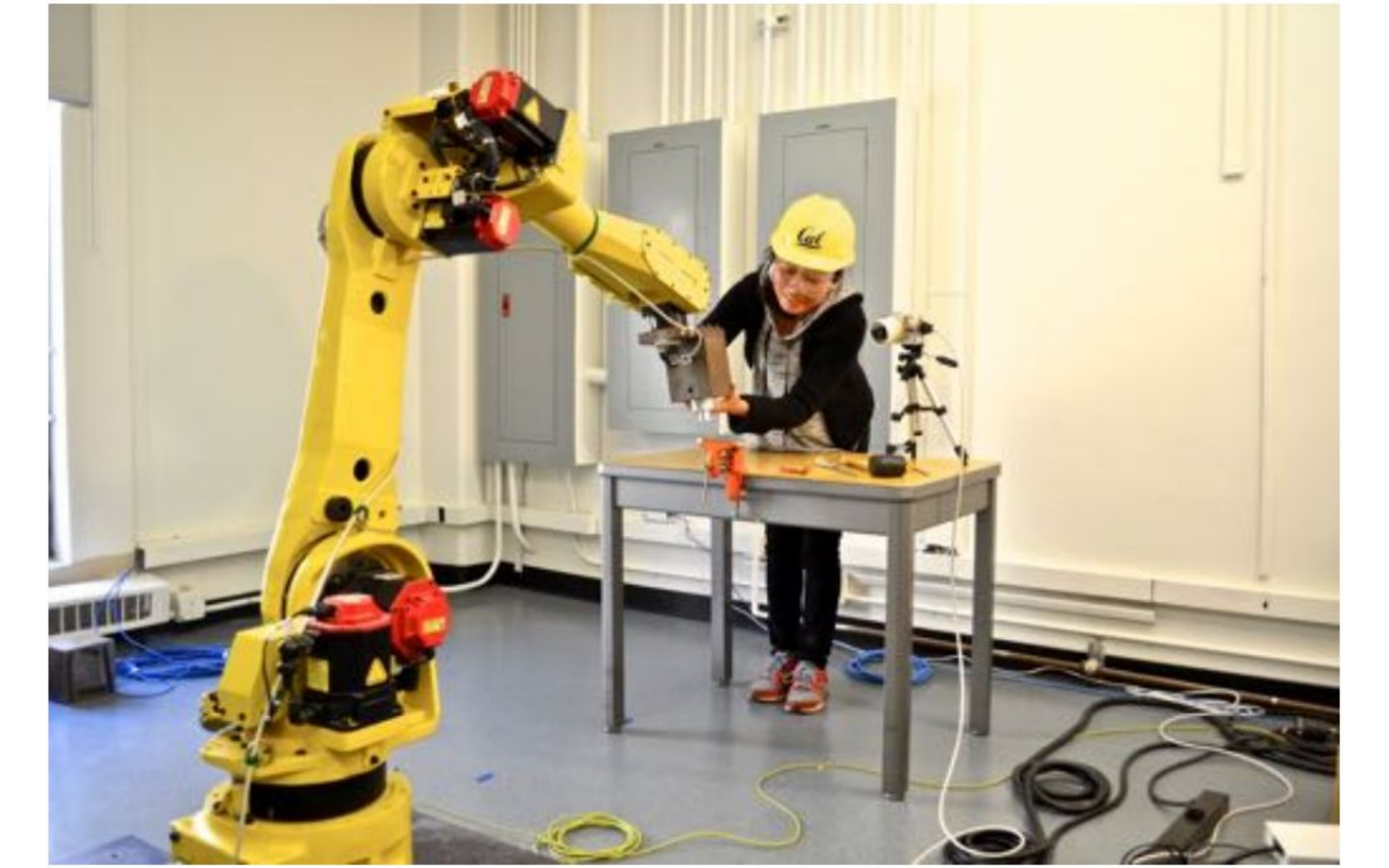}\label{fig:hri1}}\hfill
\subfigure[]{\includegraphics[width=0.48\linewidth]{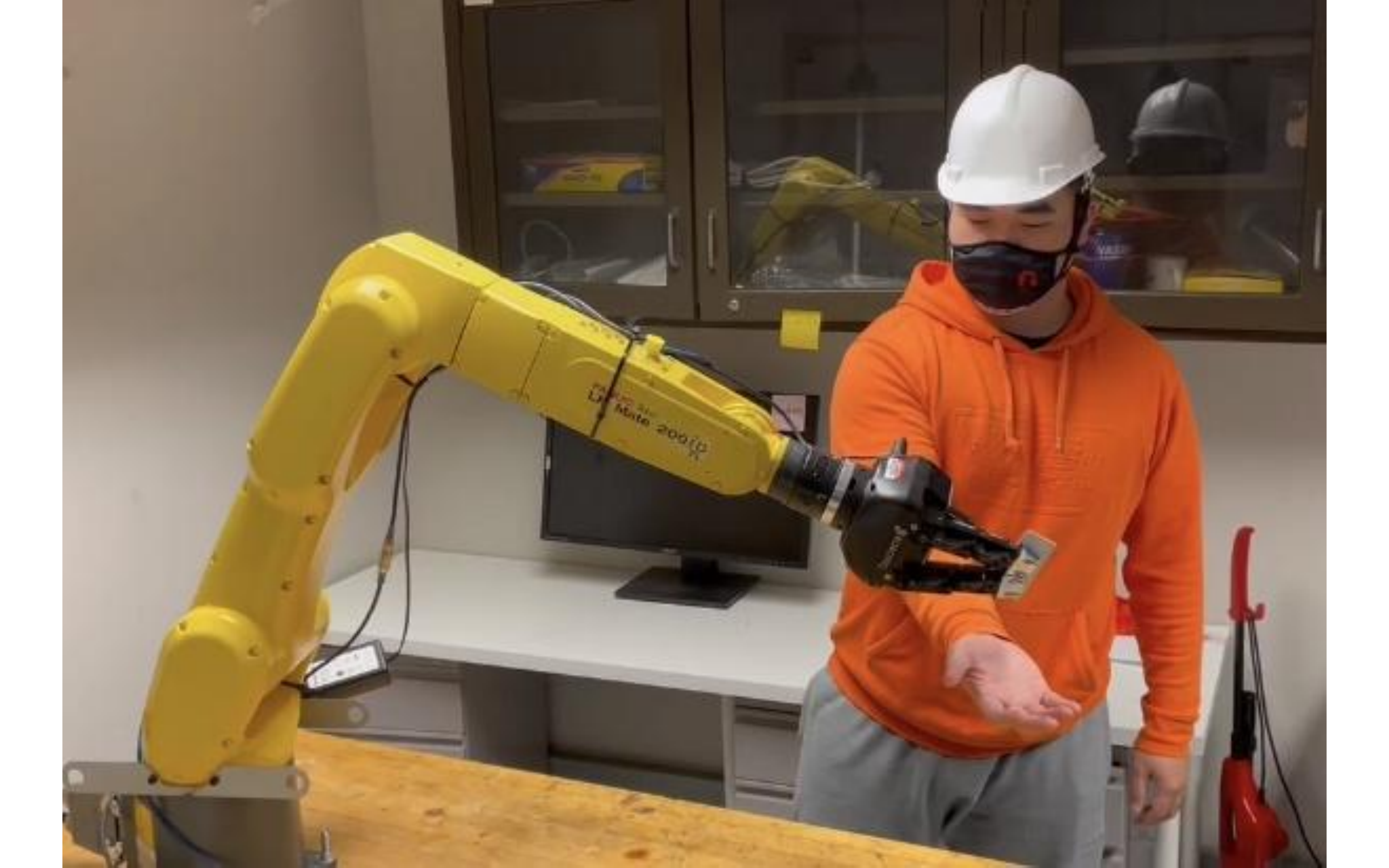}\label{fig:hri2}}
\vspace{-10pt}
    \caption{\footnotesize Examples of HRI. (a) Human-robot co-assembly \cite{7487476}. (b) Robot handover. \label{fig:hri}}
    \vspace{-15pt}
\end{figure}

Industrial robots are widely used in many applications \cite{GOPINATH2017430, 6840175}. 
The contemporary need requires robots to work intelligently in unstructured and dynamic environments, such as human-robot interaction (HRI) \cite{Christensen2021ARF, KRUGER2009628, Charalambous2013HumanautomationCI}.
\Cref{fig:hri} shows examples of HRI, where the robot co-assembles an object with the human in \cref{fig:hri1} and the robot delivers a desired workpiece to the human in \cref{fig:hri2}.

Interactive industrial robots are rarely deployed to real applications mainly due to safety concerns.
To ensure safety, most existing industrial robots either only work in highly-controlled spaces, \ie within designated work cells with fences to separate from workers \cite{Michalos2022}, or simply halt upon unexpected environmental changes, \ie human enters the work zone \cite{VILLANI2018248}.
Existing collaborative industrial robots (\ie FANUC CR series) ensure intrinsic safety by reducing the impact upon collision.
However, they are not able to actively avoid collision (interactive safety), which hinders close interaction with human workers.
Many existing works have been proposed to address the interactive safety concern \cite{1087247,6414600,7040372,9029720}.
However, the real application of those methods is difficult since most industrial robot drivers do not provide the flexibility for executing the safe control algorithms during control loops.
A jerk-bounded position control driver (JPC) \cite{jpc} was developed previously, which provides a unified interface for robot control under dynamics constraints and enables real-time trajectory modification on industrial robots.
The developed JPC enables high-frequency motion-level control and allows the integration of safe control algorithms.

To address the real-time interactive safety, the safe set algorithm (SSA) \cite{ssa} was developed. This paper investigates methods to integrate SSA into JPC so that the safe control runs in real-time within the control loop to ensure interactive safety.
However, there are several challenges.
First, a new safety index is needed to ensure control feasibility with bounded jerks (as required by JPC) while guaranteeing the robot to stay in the safe region (or forward invariance of the safe set).
Second, it is difficult to optimize hyperparameters to ensure the best performance with the new safety index.
To address the above-mentioned challenges, this paper extends SSA to a jerk-based control system, called jerk-based SSA (JSSA), and integrates JSSA to JPC.
A new safety index is introduced to accommodate the third-order system and the robot dynamics constraints.
To optimize the performance of JSSA, an extensive sensitivity analysis is conducted on the hyperparameters of JSSA.
Results show that JSSA generates provably safe control commands that
satisfy the dynamic constraints (\ie bounded jerk) in real-time and allows the industrial robot to safely interact with humans.

The contributions of this paper are the following. 1) To the best of our knowledge, this is the first paper that studies safe control algorithms (in particular, SSA) on a third-order system, while previous works focus on second-order systems \cite{ssa, 8062637,7487476}. 2) This paper integrates the JSSA to JPC \cite{jpc}, which allows the safe control algorithm to run in a real-time control loop on industrial robots. 
3) This paper studies the influence of JSSA's hyperparameters on the system performance through extensive sensitivity analysis.
4) This paper demonstrates the integrated JSSA on a FANUC LR Mate 200id/7L robot arm in real HRI tasks.

The remainder of this paper is organized as follows. 
\Cref{sec:pf} formulates the problem that JSSA addresses.
\Cref{sec:JSSA} presents the jerk-based SSA (JSSA).
\Cref{sec:exp} shows the experiment results of JSSA on the industrial robot in simulation and real HRI tasks.
Lastly, \cref{sec:conclusion} concludes the paper.

\section{Problem Formulation}\label{sec:pf}

\subsection{System Modeling}\label{sec:pf1}
\paragraph{Robot System Modeling:}
We model the dynamics of a robot arm using a discrete-time jerk control system to enforce the jerk bound.
For an n-DOF robot, the robot state in the joint space is denoted as $q=[\theta_1;~\ldots;~\theta_n;~\Dot{\theta}_1;~\ldots;~\Dot{\theta}_n;~\Ddot{\theta}_1;~\ldots;~\Ddot{\theta}_n]$.
We have the robot dynamics as 
\begin{equation}
    \label{eq:robot_joint_dynamic}
    % \begin{split}
    q_{k+1}=\begin{bmatrix}
        I_n & \tau I_n & \frac{1}{2}\tau^2 I_n\\
        0 & I_n & \tau I_n\\
        0 & 0 & I_n
        \end{bmatrix}q_k + \begin{bmatrix}
        \frac{1}{6}\tau^3 I_n\\
        \frac{1}{2}\tau^2 I_n\\
        \tau I_n
        \end{bmatrix}u_k,
    % \end{split}
\end{equation}
where $\tau$ is the sampling time. And $u=\dddot{\theta}\in U=\{u\mid u_{min}\leq u \leq u_{max}\}$, where $u_{min}\leq 0 \leq u_{max}$, denotes the bounded jerk control input. 
% The joint position, velocity and acceleration are bounded as well.
$I_n \in \mathbf{R}^{n\times n}$ is an identity matrix.
This paper considers discrete-time systems. The proposed method can be extended to continuous-time systems, which will be left for future work.

\paragraph{Environment Modeling:} 
We use $E=D\cup O$ to encode the environment, where $D=[D^1; \dots;D^{n_D}]$ denotes the states of $n_D$ dynamic agents (\ie humans), and $O=[O^1; \dots;O^{n_O}]$ denotes the states of $n_O$ static agents (\ie table).
$D^i, O^i\in \mathbf{R}^9$ encode the Cartesian position, velocity and acceleration of the agents.
This paper assumes the velocity and acceleration are bounded for all agents.
The model of the static agents is 
\begin{equation}\label{eq:static_agent_dynamics}
    \begin{split}
        O^i_{k+1}&=O^i_k.
    \end{split}
\end{equation}
And the dynamic agents are modeled as $D^i_{k+1}=f_{D^i}(D^i_k, u^{D^i}_k)$, where $u^D$ is the ``imaginary'' control of the dynamic agent $D$. $u^D$ summarizes all influencing factors (\eg the goal of the agent, the robot motion, etc), and $f_D$ is an unknown function that models the agent dynamics.
In this paper, we linearly model the dynamics as
\begin{equation}
    \label{eq:human_full_dyn}
    D_{k+1}=\underbrace{\begin{bmatrix}
        I_3 & \tau I_3 & \frac{1}{2}\tau^2 I_3\\
        0 & I_3 & \tau I_3\\
        0 & 0 & I_3
        \end{bmatrix}}_{A^{C}(\tau)}D_k + \underbrace{\begin{bmatrix}
        \frac{1}{6}\tau^3 I_3\\
        \frac{1}{2}\tau^2 I_3\\
        \tau I_3
        \end{bmatrix}}_{B^{C}(\tau)}u^D_k.
\end{equation}
Note that the system is underactuated (\ie $E$ is not always controllable by $u$).
By assuming the humans have continuous motion without sudden change, this paper adopts a constant velocity model without the loss of generality. Thus, the dynamic agent model \cref{eq:human_full_dyn} can be reduced to
\begin{equation}
    \label{eq:human_simplified_dyn}
    % \begin{split}
    D_{k+1}=A^{C}(\tau)D_k,
    % \end{split}
\end{equation}
where $D$ has the acceleration entries being zero-valued terms.
Note that it is feasible to model the full dynamics of the human behavior using a nonlinear model and identify the ``imaginary'' control online using adaptation algorithms \cite{9281312}, which will be left for future work.
With the robot state $q$ and the environment state $E$, the state of the overall system is $x=[q;~E]$.

\subsection{Safety Specification}
We use $X$ to denote the system state space. 
The safety specification requires that $x\in X$ is constrained in a closed subset $X_{S}\subseteq X$.
Note that the safety considered in this paper is collision avoidance, meaning that at any given time, there should be no collision between the robot and the agents.
An initial safety index $\phi_0(x): X\mapsto \mathbf{R}$ is specified to quantify the safety, such that $X_{S}=\{x ~|~ \phi_0(x)\leq 0\}$.
We assume $\phi_0$ is user-defined.

\subsection{Nominal Control}\label{pf3}
It is assumed that the robot system has a nominal controller such that for a given user-specified task, the nominal controller generates a jerk control sequence $u=[u_1;~u_2;\dots]$ that controls the robot to execute the task. 
Since the robot shares the environment with the dynamic agents and needs to ensure safety, the nominal control is subject to modification by JSSA.

\begin{figure}
\footnotesize
    \centering
    \includegraphics[width=\linewidth]{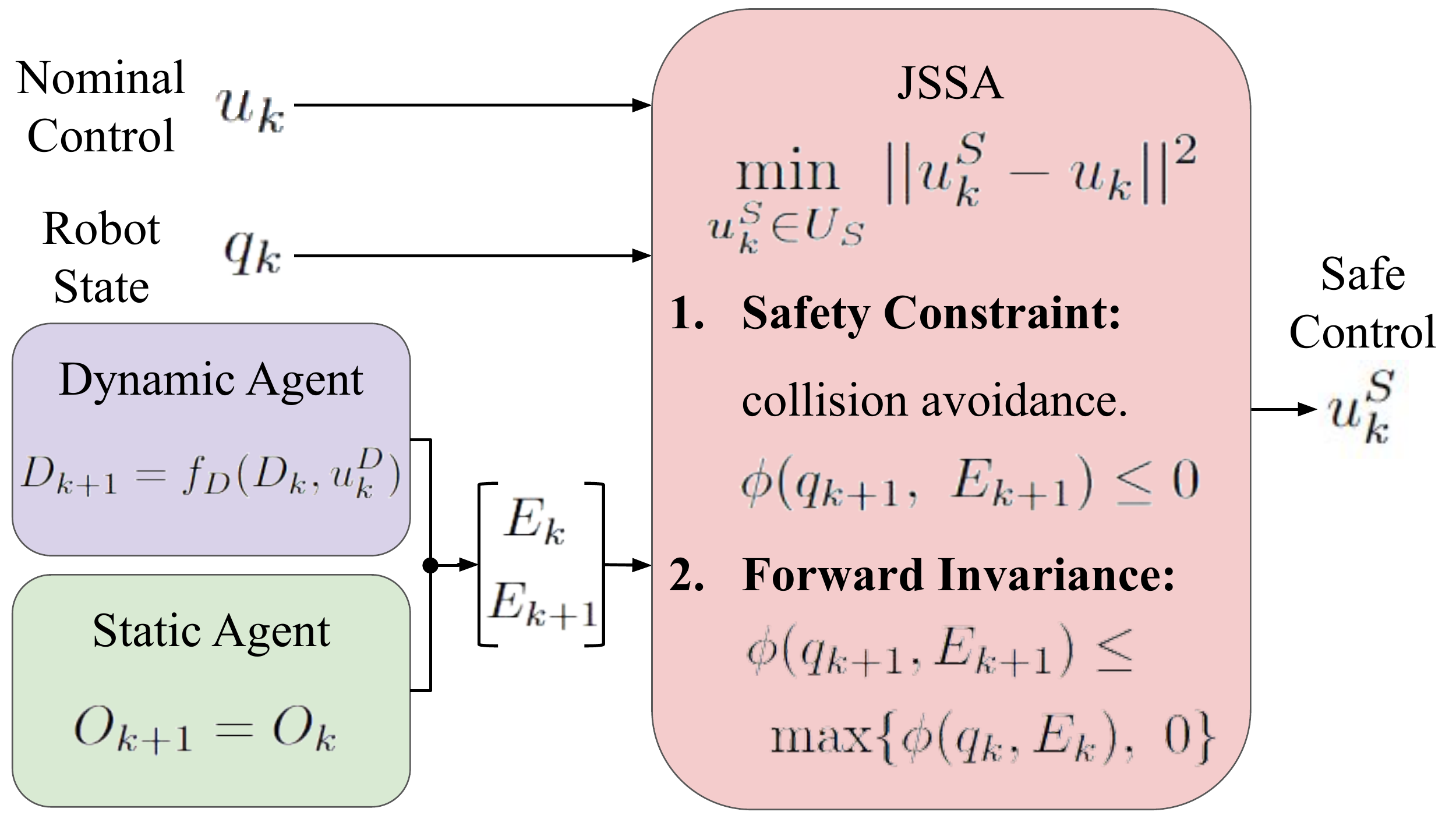}
     \vspace{-10pt}
    \caption{\footnotesize Illustration of the JSSA problem.}
    \label{fig:JSSA_problem_illustration}
    \vspace{-15pt}
\end{figure}

\subsection{Problem: Safeguard with JSSA}
The core problem is to synthesize a safeguard that monitors the nominal control and modifies it if necessary, i.e., $u_k^S = JSSA(u_k, ~q_{k},~ E_{k})$, which is also shown in \cref{fig:JSSA_problem_illustration}. The safe control $u_k^S$ should ensure forward invariance in the safe set, i.e., $x_{k+1}\in X_S$ for all $k$. In other words, the system should never leave the safe set after entering it.
A naive approach to design JSSA is to directly enforce the safety constraint $\phi_0(x_{k+1})\leq 0$ by choosing a closest $u_k^S$ to $u_k$ that satisfies the constraint. However, due to dynamic limits, it is not guaranteed that there is always a feasible control $u\in U$ that can satisfy the constraint. Hence, we need to design a new safety index $\phi$ to ensure control feasibility and forward invariance.

\section{Jerk-based Safe Set Algorithm} \label{sec:JSSA}

This section introduces the jerk-based safe set algorithm (JSSA), following the design principles of the original SSA \cite{ssa}.
The original SSA contains the following parts:
\textbf{Offline}: 1) designing the safety index $\phi$ and 2) system modeling (discussed in \cref{sec:pf1}); and
\textbf{Online}: 1) environment update and 2) real-time computation of the safe control $u^S$.
The proposed JSSA differs from other SSA-related works in that 1) this paper introduces a new safety index that is suitable for a jerk-based control system; 2) we use JSSA to safely guard JPC.

\subsection{Safety Index Synthesis}

\begin{figure}
\footnotesize
\subfigure[$X_S^*(\phi_0)$]{\includegraphics[width=0.3\linewidth]{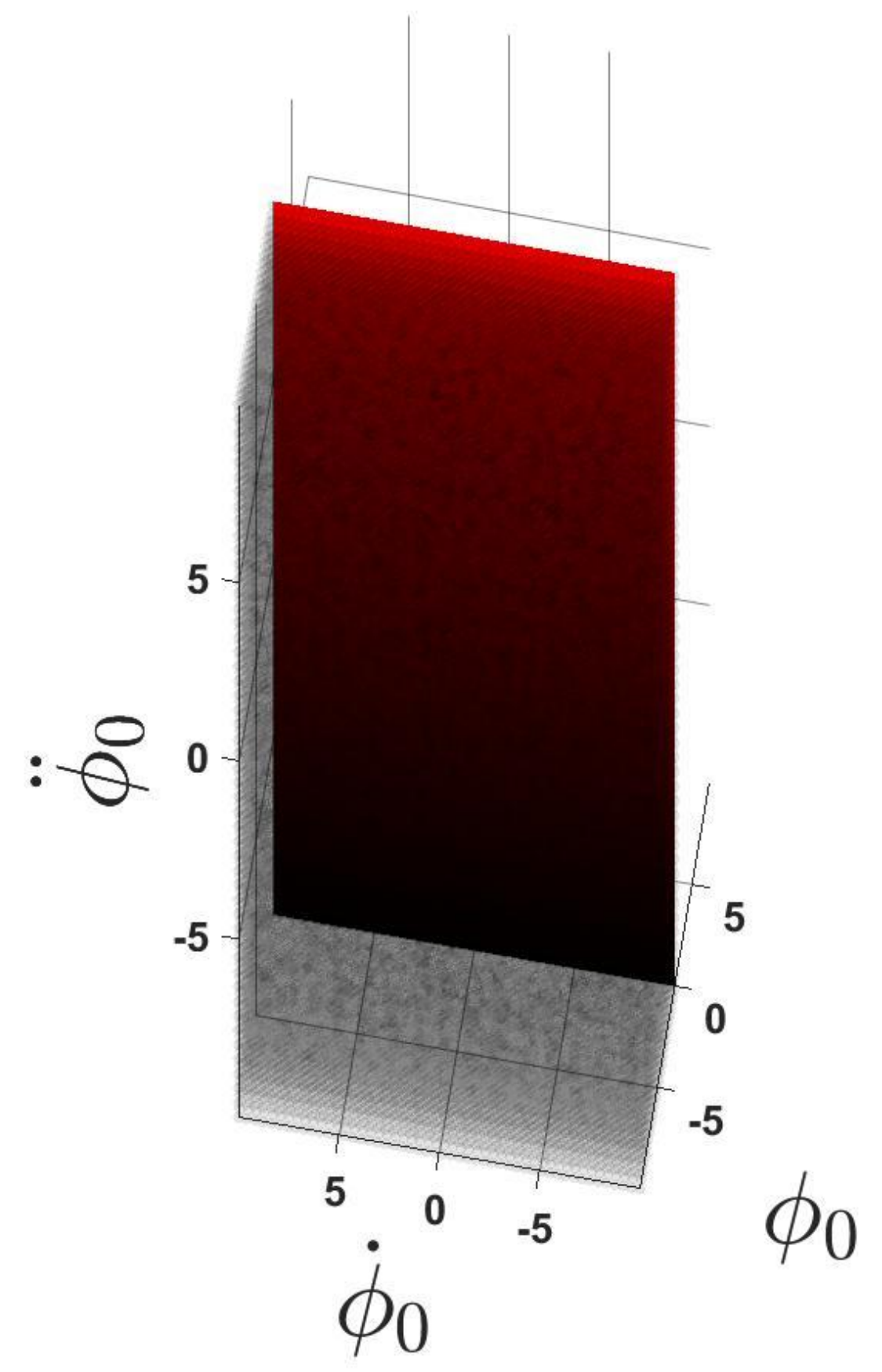}\label{fig:phase1}}\hfill
\subfigure[$X_S^*(\phi^*)$]{\includegraphics[width=0.3\linewidth]{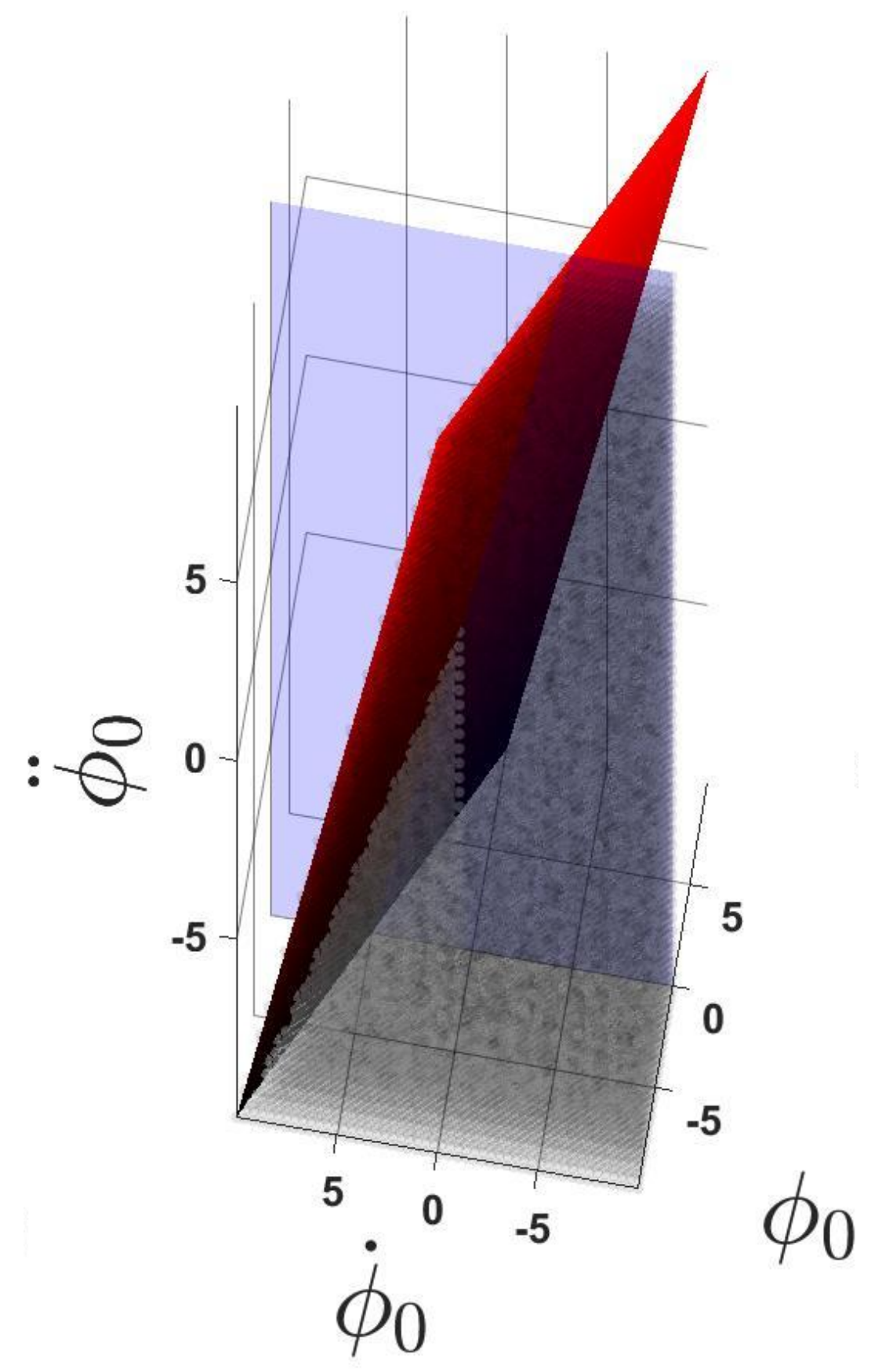}\label{fig:phase2}}\hfill
\subfigure[$X_S^*(\phi)$]{\includegraphics[width=0.3\linewidth]{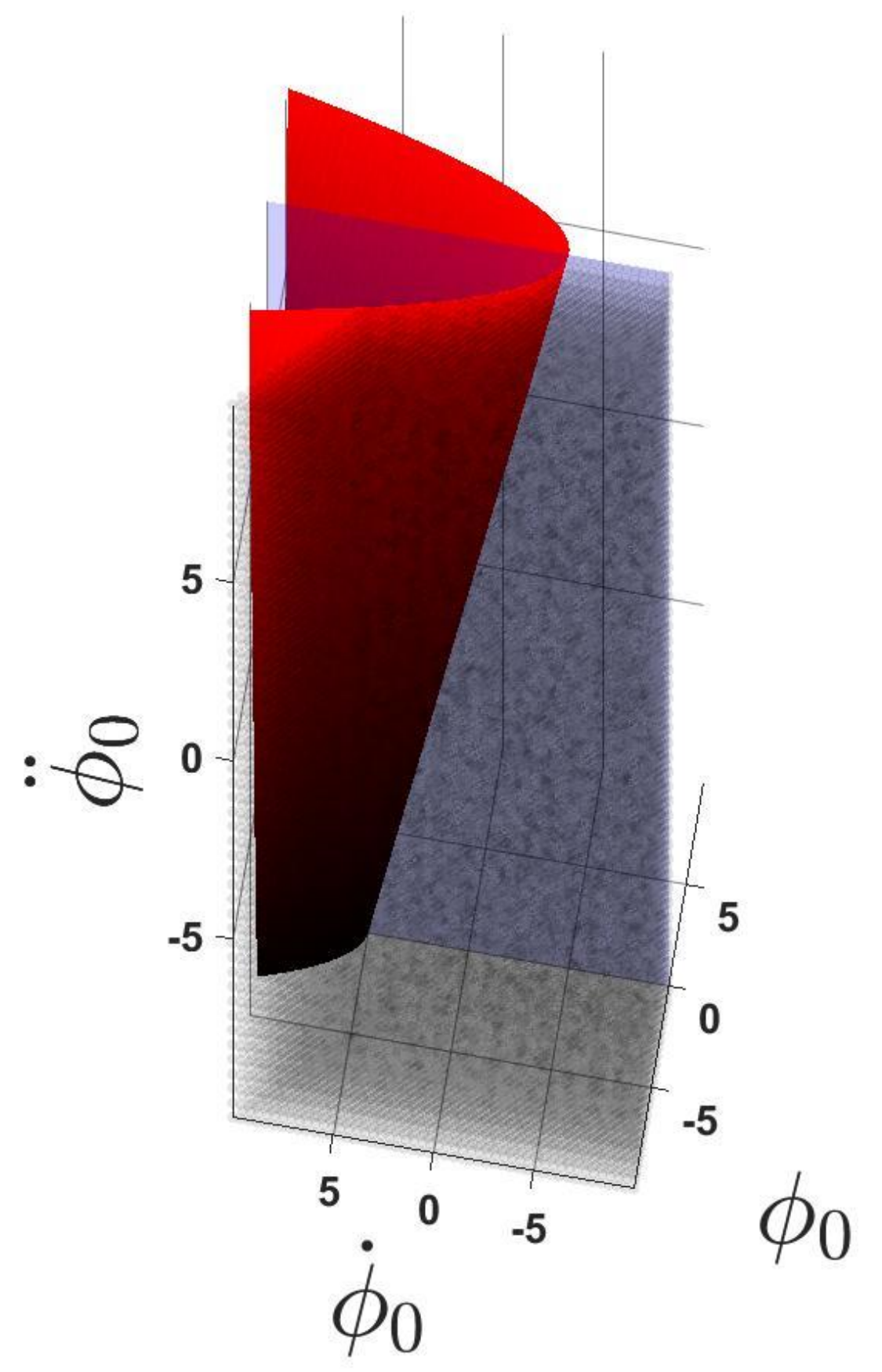}\label{fig:phase3}}
\vspace{-10pt}
    \caption{\footnotesize Relationship between $X_S^*(\phi_0)$, $X_S^*(\phi^*)$, and $X_S^*(\phi)$. \label{fig:phase_portrait}}
    \vspace{-15pt}
\end{figure}

This paper considers the safety specification as collision avoidance in the 3D Cartesian space, where $\phi_0=d_{min}-d$ and the safe set is defined as $X_S=\{x=[q;E]\mid \phi_0(x)\leq 0\}$ (the shaded area in \cref{fig:phase1}).
The constant $d_{min}$ is the safety margin, whereas $d$ is the real-time minimum distance between the robot and the agents, which can be computed using capsule representations \cite{7487476}.
Note that this paper considers only the minimum distance between the robot and all agents, which indicates that the safeguard is designed to avoid imminent collision.
This strategy works under the assumption that there is only one critical obstacle at a given time (sparse obstacle environment), which is realizable in a typical HRI setting.
Consideration of multiple distance constraints will be left for future work.

Given the safety specification, our goal is to synthesize a safety index $\phi$ to satisfy the following two conditions. 
%1) here exists a nontrivial subset $Q_S^*\subseteq Q_S$ that is \textbf{forward invariant} (\ie $Q_S^*=\{q~|~\phi(q,E)\leq 0\}\subseteq \{q~|~\phi_0(q,E)\leq 0\}~ \forall E$) and 
1) There always exists a \textbf{feasible} control input to keep $\phi(x_{k+1})\leq 0$ for any state that $\phi(x_{k})\leq ~0$ 
\begin{equation}\label{eq:safe_control_set}
    U_S(x_k)=\{u^S\in U \mid \phi(x_{k+1})\leq ~0\}\neq \emptyset,
\end{equation}
where $U_S$ is the set of safe control. 
2) By always choosing the control in the set $U_S$, \textbf{forward invariance} within $X_S$ is guaranteed, i.e., $X_S^*(\phi)\subseteq X_S$, where $X_S^*(\phi)$ is the set of reachable states that start from $X_0=\{x\in X_S\mid \dot\phi_0=\ddot\phi_0=\ldots=\phi_0^{(m-1)}=0\}$ with the control input selected from $U_S$ in \eqref{eq:safe_control_set}. And $m$ is the relative degree from $\phi_0$ to $u$. 

To meet the second requirement, we directly leverage the theorem in \cite{ssa}, which proved that if 1) the safety index is defined as $\phi_m = \phi_0^* + \lambda_1\dot \phi_0 +\lambda_2\ddot \phi_0 +\ldots + \lambda_{m-1}\phi_0^{(m-1)}$, where $\phi_0^*$ defines the same set as $\phi_0$ ($\{x\mid\phi_0\leq 0\}\equiv \{x\mid\phi_0^*\leq 0\}$); and all roots of $1+\lambda_1 s+\lambda_2 s^2 +\ldots + \lambda_{m-1}s^{m-1}=0$ are negative real; and 2) there is always a feasible control to realize the control strategy $\dot\phi\leq0$ when $\phi=0$, then the set $X_S^*(\phi_m)$ is forward invariant inside $X_S$. Note these results are proved in continuous time. In the following discussion, we ignore the continuous-time to discrete-time gap in the derivation, which is valid when the sampling time is sufficiently small. The formal analysis with discrete-time systems will be left for future work.

Since \cref{eq:robot_joint_dynamic} defines a jerk control system, then $\phi$ should include the first and second derivatives of $\phi_0$. 
Moreover, we introduce $\phi_0^*=d_{min}^2-d^2$ to nonlinearly shape the gradient $\phi$ at the boundary of the safe set. 
Therefore, the safety index $\phi$ is parameterized as
\begin{equation}\label{eq:SI}
\begin{split}
    \phi =& \phi_0^*+\lambda_1\dot{\phi}_0+\lambda_2\ddot{\phi}_0,\\
    =&d^2_{min}-d^2-\lambda_1\Dot{d}-\lambda_2\Ddot{d},
\end{split}
\end{equation}
where $\lambda_1, \lambda_2 \in \mathbf{R}$ are tunable coefficients, and all roots of $1+\lambda_1s+\lambda_2s^2=0$ are negative real. To meet the first requirement that $U_S$ is non-empty, we need to impose additional constraints on $\lambda_1$ and $\lambda_2$ to ensure that when $\phi =0$, $\dot\phi <0$. Since $\dot\phi = -2d\dot{d}-\lambda_1\ddot{d}-\lambda_2\dddot{d}$ and $\dddot{d}$ depends on the control $u$, we just need to choose $\lambda_1$ and $\lambda_2$ such that the following inequality holds:
\begin{equation}\label{eq:minimax}
\begin{split}
    \max_{x\text{ s.t. }\phi(x)=0}\min_{u\in U}-2d\dot{d}-\lambda_1\ddot{d}-\lambda_2\dddot{d}(u)\leq& 0.
\end{split}
\end{equation}
This paper empirically verifies the design by solving the minimax problem approximately using samples. As a future work, we will derive explicit conditions on $\lambda_1$ and $\lambda_2$ from \cref{eq:minimax}.

\cref{fig:phase_portrait} illustrates the possible forward invariant set under different designs of the safety index: a) $\phi_0$, b) $\phi^*=\phi_0+\lambda_1\dot\phi_0+\lambda_2\ddot\phi_0$, and c) $\phi$ in \eqref{eq:SI}. These sets ignores the dynamic constraints by assuming $U_S$ is always non-empty, hence may not be ``true'' forward invariant sets. 
The red surfaces in \cref{fig:phase_portrait} indicate the corresponding safety index equals 0, where the red intensity on each surface increases as $\ddot{\phi}_0$ increases. The blue transparent surfaces are $\phi_0=0$. The shaded area shows the possible forward invariant set $X_S^*(\phi_0)$, $X_S^*(\phi^*)$ and $X_S^*(\phi)$. 
In \cref{fig:phase3}, we introduce nonlinearity to the safety index by substituting $\phi_0$ with $\phi_0^*$. The shaded area $X_S^*(\phi)$ hence has a nonlinear boundary.
Note that the shaded forward invariant set in the nonlinear case only shows a subset of the maximal forward invariant set, while the exact form of $X_S^*(\phi)$ is left for future work.

\subsection{Real-Time Control Synthesis}
Given the offline designed safety index $\phi$ and the system model, this section derives the real-time safe control $u^S$.
To ensure safety, $u^S\in U_S$ \eqref{eq:safe_control_set}.
Due to the complex geometries of the robots and the agents, this paper uses the minimum distance between the robot and all agents (critical point pair). 
Let $M=[p_R;~\dot{p}_R;~\ddot{p}_R;]$, $H=[p_H;~\dot{p}_H;~\ddot{p}_H] \in \mathbf{R}^9$ be the critical point pair states on the robot and the agents. Note that $H$ can be a point either on $O$ or $D$. The dynamics of $M$ can be written as
\begin{equation}
    \label{eq:robot_cart_dyn}
    M_{k+1}=A^{C}(\tau)M_k+B^{C}(\tau)j_k,
    % \end{split}
\end{equation}
where $j=\dddot{p}_R$ is the jerk control input in the Cartesian space.
And the dynamics of $H$ follows either \cref{eq:static_agent_dynamics} or \cref{eq:human_simplified_dyn}.
Define the relative state between the critical point pair as
\begin{equation}\label{eq:state_diff}
    \begin{split}
        \delta_{k+1} &= M_{k+1}-H_{k+1}\\
        &=A^{C}(\tau)M_k+B^{C}(\tau)j_k-H_{k+1}
    \end{split}
\end{equation}
Note that the jerk control model \cref{eq:robot_joint_dynamic} is in the joint space and \cref{eq:state_diff} has the control input $j$ in the Cartesian space. 
Let $J\in \mathbf{R}^{3\times n}$ be the jacobian matrix of the n-DOF robot without the rotation entries,
we have $\dot{p}_R=J\dot{\theta}$.
And thus, to relate $u$ and $j$, we have
\begin{equation}\label{eq:j2u}
    j_k=\ddot{J}\cdot[0,~I_n,~0]\cdot q_k+2\cdot\dot{J}\cdot[0,~0,~I_n]\cdot q_k+J\cdot u_k.
\end{equation}
We can then express \cref{eq:state_diff} in terms of $u$ as
\begin{equation}
    \begin{split}
        \delta_{k+1}&= \Delta+B^C(\tau)\cdot J \cdot u_k,\\
        \Delta&=A^C(\tau)M_k+B^C(\tau)\cdot(\ddot{J}\cdot\dot{\theta}_k+2\dot{J} \cdot\ddot{\theta}_k)-H_{k+1}.
    \end{split}
\end{equation}
And the safety components can be calculated as
\begin{equation}
    \begin{split}
        d^2=&\delta^T\underbrace{\begin{bmatrix}
        I_3 & 0 & 0\\
        0 & 0 & 0\\
        0 & 0 & 0
        \end{bmatrix}}_{U_1} \delta,~~~
        \Dot{d}=\frac{1}{d}\delta^T\underbrace{\begin{bmatrix}
        0 & I_3 & 0\\
        0 & 0 & 0\\
        0 & 0 & 0
        \end{bmatrix}}_{U_2} \delta,\\
        \Ddot{d}=&-\frac{\dot{d}^2}{d}+\frac{1}{d}\delta^T\underbrace{\begin{bmatrix}
        0 & 0 & 0\\
        0 & I_3 & 0\\
        0 & 0 & 0\end{bmatrix}}_{U_3}\delta+\frac{1}{d}\delta^T\underbrace{\begin{bmatrix}
        0 & 0 & I_3\\
        0 & 0 & 0\\
        0 & 0 & 0
        \end{bmatrix}}_{U_4} \delta.
    \end{split}
\end{equation}
Therefore, the constraint $\phi(x_{k+1})\leq 0$ can be translated to 
\begin{equation}\label{eq:safety_condition}
    \begin{split}
        0\geq \phi d=&d^2_{min}\sqrt{\delta^TU_1\delta}-(\delta^TU_1\delta)^{\frac{3}{2}}-\lambda_1\delta^TU_2\delta-\\
        &\lambda_2(-(\frac{\delta^TU_2\delta}{\sqrt{\delta^TU_1\delta}})^2+\delta^TU_3\delta+\delta^TU_4\delta),\\
        \approx&d^2_{min}\sqrt{\Delta^TU_1\Delta}-(\Delta^TU_1\Delta)^{\frac{3}{2}}-\lambda_1\Delta^TU_2\Delta+\\
        &\lambda_2(\frac{(\Delta^TU_2\Delta)^2}{\Delta^TU_1\Delta}-\Delta^TU_3\Delta-\Delta^TU_4\Delta)\\
        &-\underbrace{2(\lambda_1\Delta^TU_2+\lambda_2\Delta^TU_3+\lambda_2\Delta^TU_4)B^C(\tau)J}_{L}u\\
        =& S-Lu,
    \end{split}
\end{equation}
where the higher-ordered $u$ terms are eliminated for approximation due to small $\tau$. 
Thus, the safe control $u^S$ is computed by projecting the nominal control to the safe control set. The $u^S$ can be solved by the following optimization
\begin{equation}\label{eq:SSA_optimization}
    \begin{split}
        \min_{u_k^S}&(u_k^S-u_k)^T V(u_k^S-u_k),\\
        &\st L u_k^S\geq S,~u_k^S\in U.
    \end{split}
\end{equation}
where $V\in \mathbf{R}^{n\times n}$ is a positive definite cost matrix. 
\begin{comment}
The solution to \cref{eq:SSA_optimization} can be summarized as  
\begin{enumerate}
    \item If $L u_k\geq S$, $u_k^S=u_k$. The system is safe and the nominal control will maintain the system to be safe.
    \item If $L u_k < S$, by solving \cref{eq:SSA_optimization} we have 
    \begin{equation}
        \label{eq:u_s}
        u_k^S=u_k-\frac{(Lu_k-S)V^{-1}L^T}{LV^{-1}L^T},
    \end{equation}
\end{enumerate}
\end{comment}

\subsection{JSSA Integration to JPC}
This paper integrates the JSSA to the JPC to safely guard the nominal control generated by JPC.
Given a user-specified task $P = [P_1; P_2; \ldots; P_{N}]$, a sequence of $N$ joint trajectory points with sampling time $T$, the JPC generates an open-loop jerk control $u=[u_1;~u_2;~\dots;~u_{(N\cdot T)/\tau}]$ to track $P$.
The JSSA checks if the nominal control is safe according to \cref{eq:safety_condition} and modifies it using \eqref{eq:SSA_optimization} if necessary.

\begin{figure}
\footnotesize
    \centering
    \includegraphics[width=\linewidth]{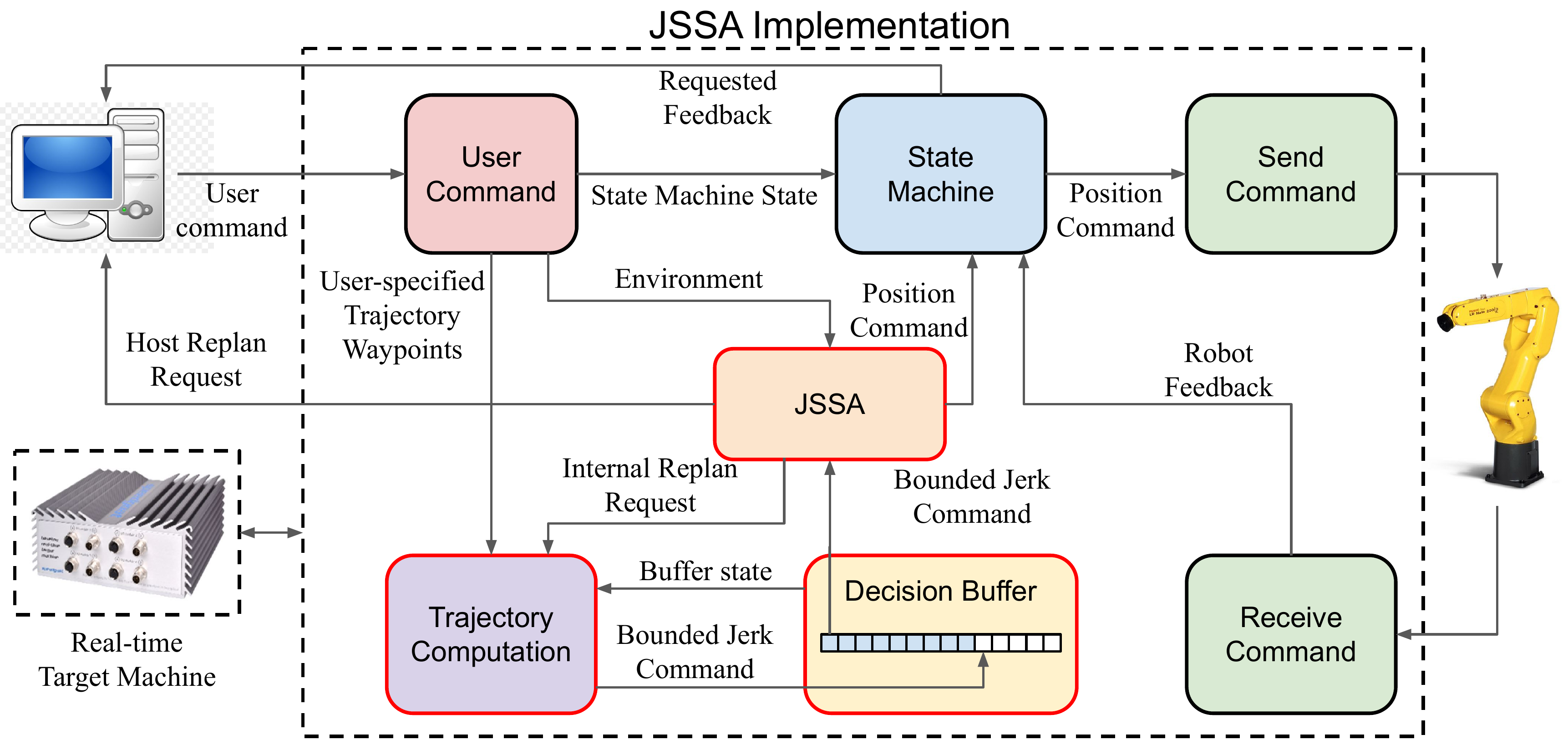}
    \caption{\footnotesize JSSA Implementation Diagram. The nominal controller is the JPC \cite{jpc} (purple) and the JSSA is integrated (orange). The integrated controller is deployed to the Speedgoat real-time target machine, which is connected in between of the host and the robot. Black-bounded modules: 1kHz. Red-bounded modules: 125Hz.
    \label{fig:JSSA_integration}}
    \vspace{-15pt}
\end{figure}

\cref{fig:JSSA_integration} shows the implementation diagram of the integration.
The JSSA module is connected to the downstream of the buffer that stores the nominal $u$.
The JSSA module internally tracks the robot state, takes environment measurements and the nominal control, then outputs the safe control $u^S$.
The position control signal is integrated after JSSA to ensure the jerk bound constraint on final commands.

Note that the JPC is an open-loop controller. After JSSA modifies the pre-computed buffer output, the remaining buffer would deviate from the original task $P$.
Meanwhile, simply commanding the robot back to the trajectory might violate the jerk bound constraint.
Therefore, JSSA adopts a replan mechanism to get back on track as shown in \cref{fig:JSSA_integration}.
It sends a replan request to the host after JSSA turns inactive (\ie robot back to safe set) so that the host plans a new user task and JPC generates a new nominal control sequence that ensures bounded jerk and resumes the original task.
Notably, the host replan time is unknown to the controller. Thus, the JSSA also sends an internal replan request to stabilize the robot once the robot is safe and avoids drifting away during the time waiting for the new task from the host.

% \section{Hardware Implementation} \label{sec:integration}

\begin{figure}
\footnotesize
\subfigure[]{\includegraphics[width=0.32\linewidth]{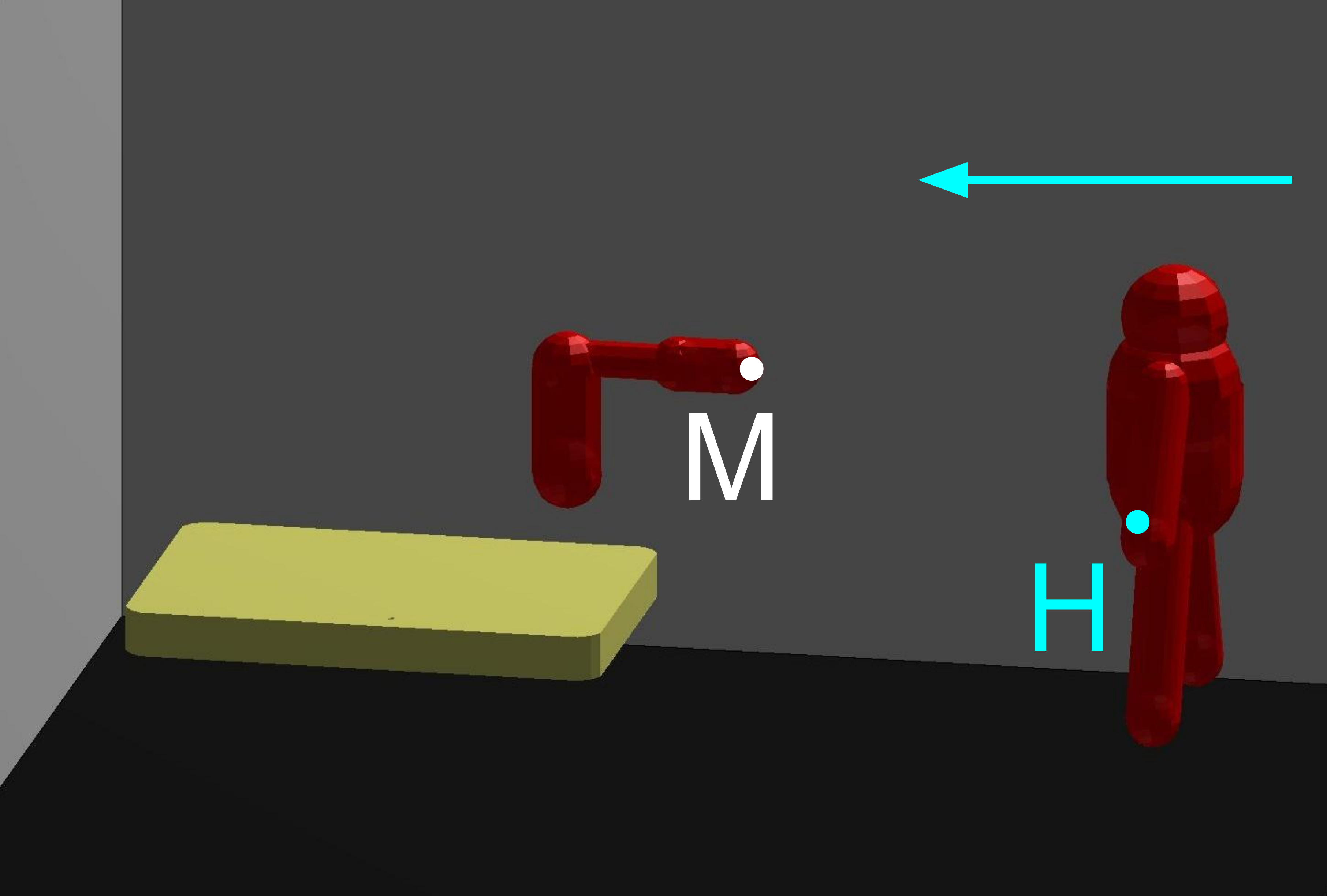}\label{fig:sim_env}}\hfill
\subfigure[]{\includegraphics[width=0.32\linewidth]{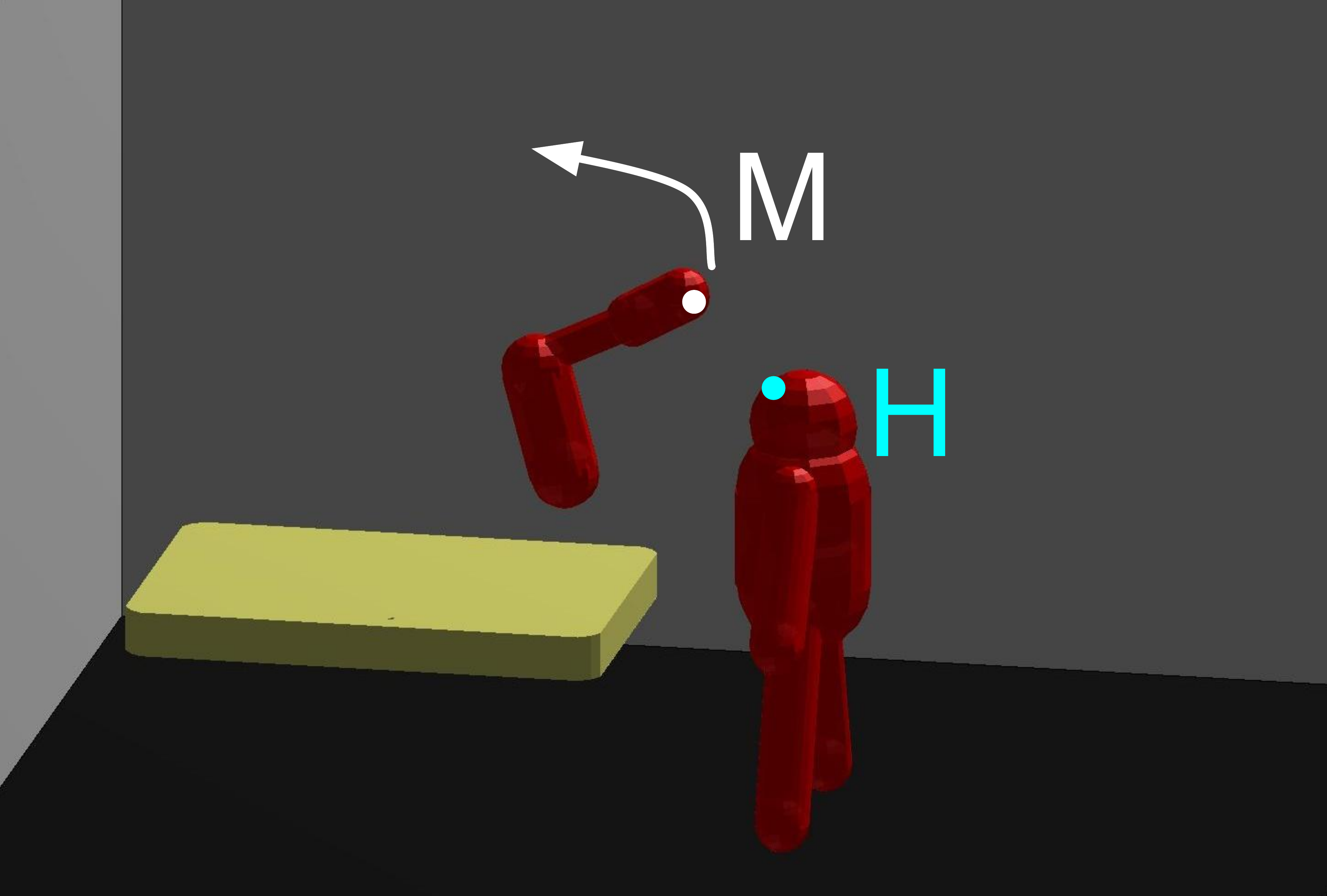}\label{fig:sim_ca}}\hfill
\subfigure[]{\includegraphics[width=0.32\linewidth]{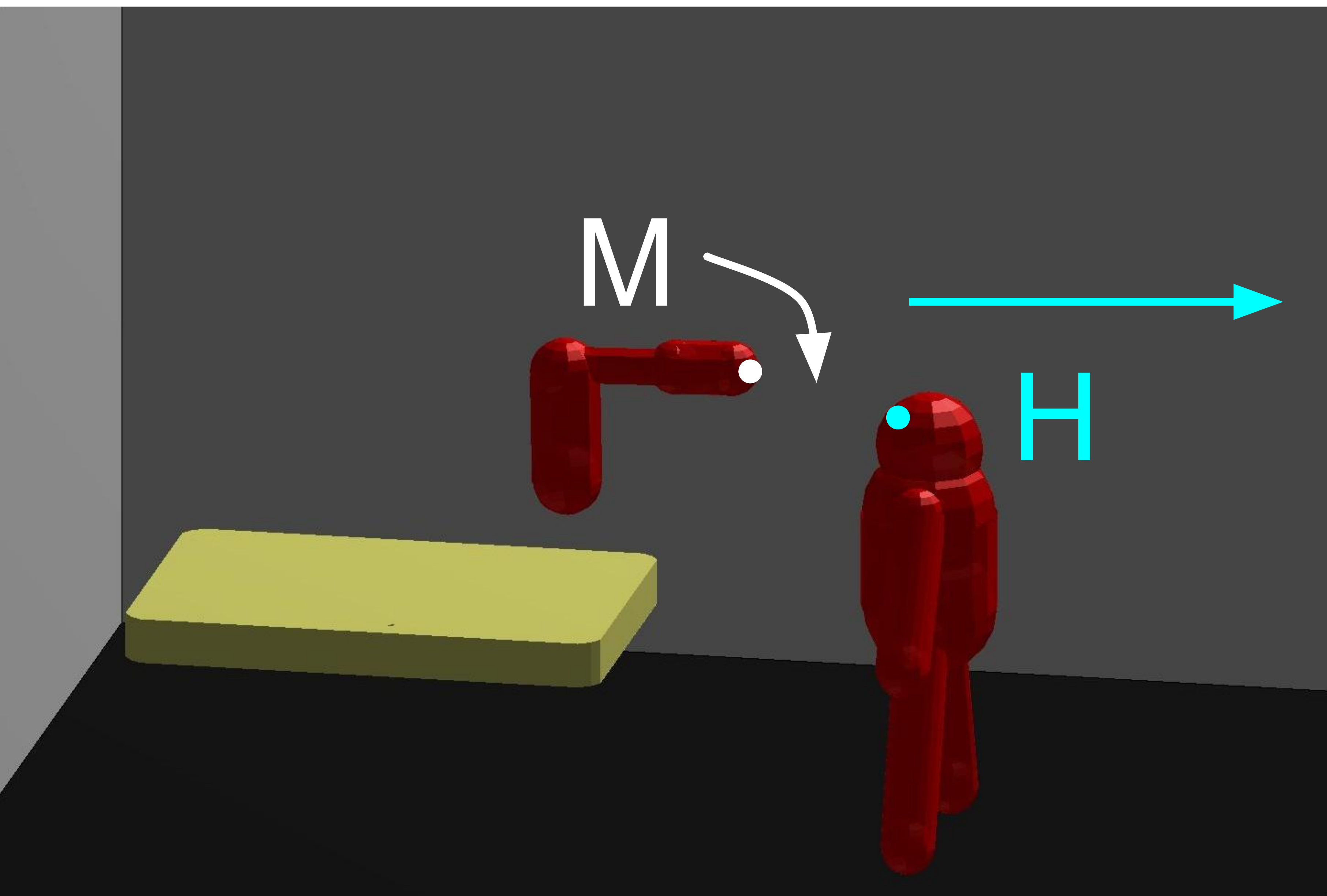}\label{fig:sim_back_to_task}}
\vspace{-10pt}
    \caption{\footnotesize Visualization of JSSA collision avoidance in simulation. $M$ and $H$ indicate the critical point pair. The arrows indicate the motion of the robot and the human. (a) The human moves toward the robot. (b) The robot avoids collision when the human gets too close. (c) The robot continues the task (stay at home) when the human leaves. \label{fig:sim_collision_avoidance}}
    \vspace{-15pt}
\end{figure}

\section{Experiment Results} \label{sec:exp}
In this section, we study the performance of JSSA on the FANUC LR Mate 200id/7L robot, a 6-DOF industrial robot, in both simulation and real human-robot interaction. 
The FANUC robot provides a position control interface via Ethernet (called \textit{stream motion}), which requires a 125Hz position control sequence, $\tau=\SI{0.008}{\second}$, with bounded jerk. 
In addition, according to our practical experience, the position control interface requires a 1kHz communication for stable performance.
Similar to the hardware implementation in \cite{jpc}, we have a Speedgoat baseline real-time target machine (SG) in between as shown in \cref{fig:JSSA_integration}. The SG and the robot have stable Ethernet communication at 1kHz.
The JSSA is integrated to the JPC in the real-time control loop, which is deployed to the SG.
In the following discussion, the implementation has the jerk bound for each joint set to $\pm[3798;3408;3505;7011;7011;10712]^\circ/s^3$, $\lambda_1=3$, $\lambda_2=1$, $V=I_6$, $d_{min}=0.05m$. By sampling in the state space, we have numerically verified the chosen $\lambda_1$ and $\lambda_2$ in the experiments satisfy the contraint in \cref{eq:minimax}.

\begin{figure}
\footnotesize
\centering
\input{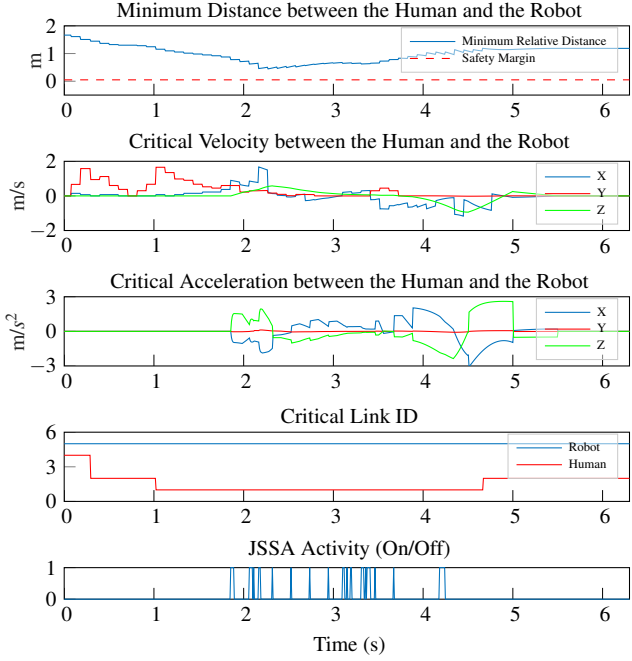}
\vspace{-10pt}
\caption{\footnotesize The simulation profile of the collision avoidance performance of JSSA ($\lambda_1=3$, $\lambda_2=1$) on the FANUC robot. Human link ID: 1) Head. 2) Core body. 3) Right arm. 4) Left arm. 5) Right leg. 6) Left leg. Robot link ID corresponds to the link number from the base to the end-effector.}
    \label{fig:sim_ca_profile}
\vspace{-15pt}
\end{figure}

\subsection{Collision Avoidance Simulation}
We test the JSSA for collision avoidance in simulation. \Cref{fig:sim_collision_avoidance} shows the simulation environment. We use capsules to simplify the geometries of the robot (5 capsules) and the human (6 capsules). 
$M$ and $H$ indicate the critical point pair. The arrows indicate the motion of the robot and the human.
The robot task is to stay at home position. 
The human only has motion in the $XY$ plane, which is generated by the human moving the mouse.
Initially, the human is far away from the robot, and thus, the robot stays idle. Then, the human starts approaching in front of the robot as shown in \cref{fig:sim_env}.
As the human gets closer, JSSA detects the potential collision and modifies the nominal control to lift the robot up to avoid collisions (\cref{fig:sim_ca}).
As the human moves away from the robot (\cref{fig:sim_back_to_task}), the robot task becomes feasible and safe. Thus, the robot returns to the home position.

\Cref{fig:sim_ca_profile} illustrates the detailed profile of the collision avoidance process visualized in \cref{fig:sim_collision_avoidance}.
The first plot demonstrates that the JSSA consistently maintains a minimum distance larger than the safety margin to avoid collision with the human.
The closest distance the human reaches is $\SI{0.446}{\meter}$.
The second and the third plot displays the relative velocity and acceleration profiles at the critical point pair.
Note that the velocity is not smooth since the sampled human velocity is not smooth.
The fourth plot indicates the critical link id of the robot and the human. The critical link is the link that contains the critical point.
We can see for the robot, the critical link remains to be 5 (end-effector) throughout the task because the human approaches in front of the robot.
For the human, the critical link is initially 4 (left arm), then 2 (core body), then 1 (head), and then 2 as the human steps back.
The last plot shows the JSSA activity.
We observe that the JSSA is triggered at \SI{1.856}{\second} and ends at \SI{4.240}{\second}.
When the JSSA is triggered, there is still a decent margin in the minimum distance as shown in the first plot. This is because the relative velocity and acceleration trigger the JSSA to actively avoid potential collisions.
The safe jerk control remains approximately in the range of $\pm2000^\circ/s^3$ for all joints, which is bounded within the jerk limit.

% \begin{figure}
% \footnotesize
% \centering
% \input{fig/simulation/k1_8_k2_6}
% % \input{fig/simulation/tracking_base}
% \vspace{-10pt}
% \caption{The simulation profile of the collision avoidance performance of JSSA ($\lambda_1=8$, $\lambda_2=6$) on the FANUC robot.}
%     \label{fig:sim_ca_profile_2}
% \vspace{-15pt}
% \end{figure}

\subsection{Safety Index Parameter Tuning}

\begin{table*}
\small
\centering
\begin{tabular}{c  c  |  c  c  c} 
\toprule
 &  & \multicolumn{3}{c}{$\lambda_1$}\\
 &  & 6 & 7 & 8\\  
\midrule
\multirow{3}{*}{$\lambda_2$} 
   & 6 & (0.317; 1.848; 4.832; 0.536; 0.640; 0.304) & (0.343; 0.080; 4.848; 0.760; 0.654; 0.382) & (0.382; 0.080; 5.024; 1.064; 0.639; 0.413)\\
   & 7 & (0.284; 1.848; 4.544; 0.688; 0.665; 0.305) & (0.313; 0.080; 4.840; 0.864; 0.649; 0.312) & (0.342; 0.080; 4.840; 0.960; 0.641; 0.345)\\
   & 8 & (0.284; 1.856; 4.816; 0.736; 0.631; 0.303) & (0.291; 1.856; 4.840; 0.776; 0.643; 0.261) & (0.316; 0.080; 5.024; 0.944; 0.633; 0.299)\\
\bottomrule
\end{tabular}
\caption{\footnotesize Comparison of the JSSA performance with different $\lambda_1$ and $\lambda_2$. Each entry includes (minimum relative distance ($m$), JSSA first trigger time ($s$), JSSA last trigger time ($s$), JSSA duration ($s$), average critical velocity ($m/s$), average critical acceleration ($m/s^2$)). \label{table:sensitivity}}
\vspace{-15pt}
\end{table*}

This section analyzes the effect of the hyperparameters in $\phi$ (\ie $\lambda_1$ and $\lambda_2$).
We study the system performance when $\lambda_1, \lambda_2\in\{6,7,8\}$, with the same human trajectory in \cref{fig:sim_collision_avoidance}.
\Cref{table:sensitivity} demonstrates an extensive sensitivity analysis. Each entry includes 1) the minimum relative distance, 2) JSSA first trigger time, 3) JSSA last trigger time, 4) JSSA active duration, 5) mean critical velocity, and 6) mean critical acceleration.

\begin{figure}
\footnotesize
\centering
\subfigure[Safety index with increased $\lambda_1$.]{\includegraphics[width=0.48\linewidth]{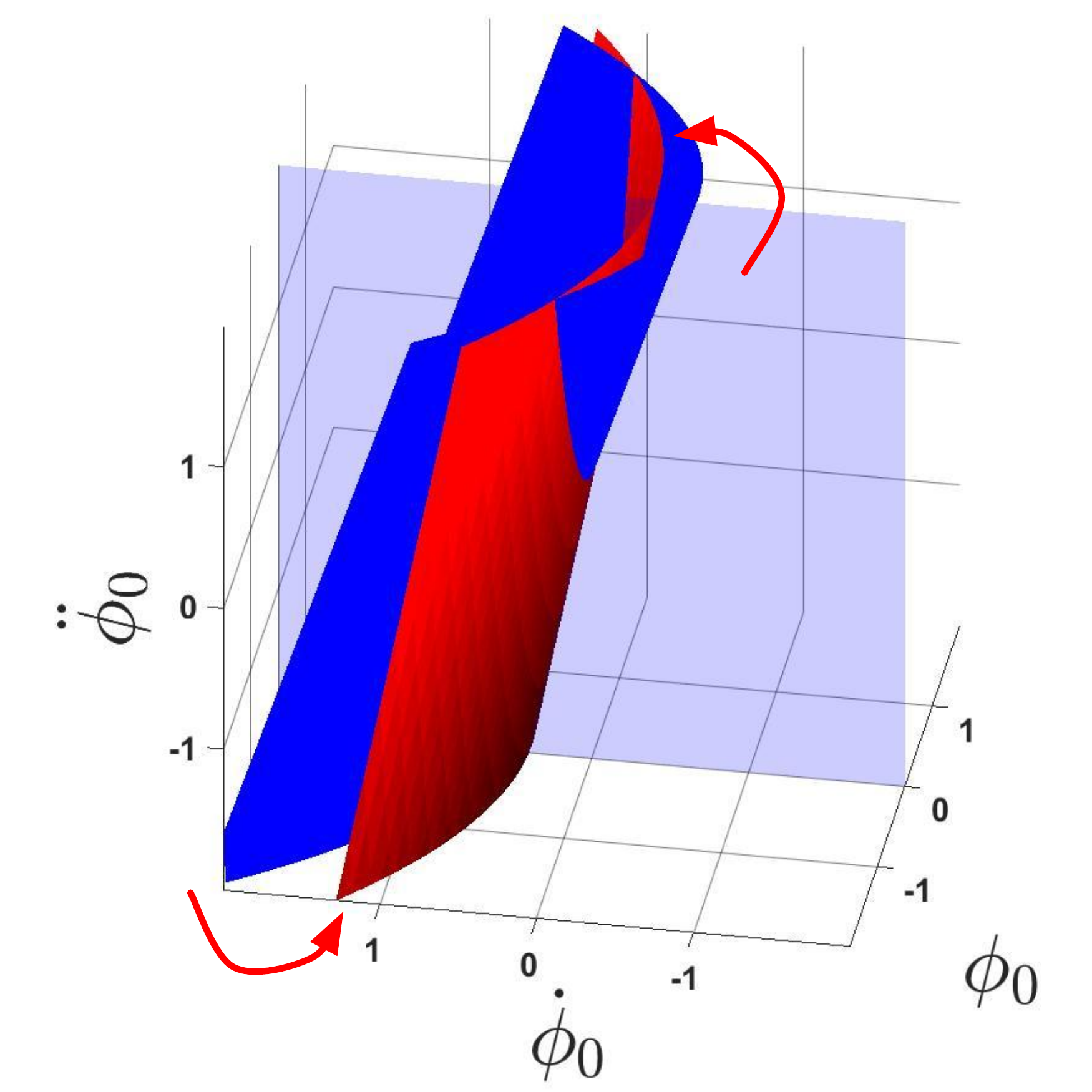}\label{fig:phase_change1}}\hfill
\subfigure[Safety index with increased $\lambda_2$.]{\includegraphics[width=0.48\linewidth]{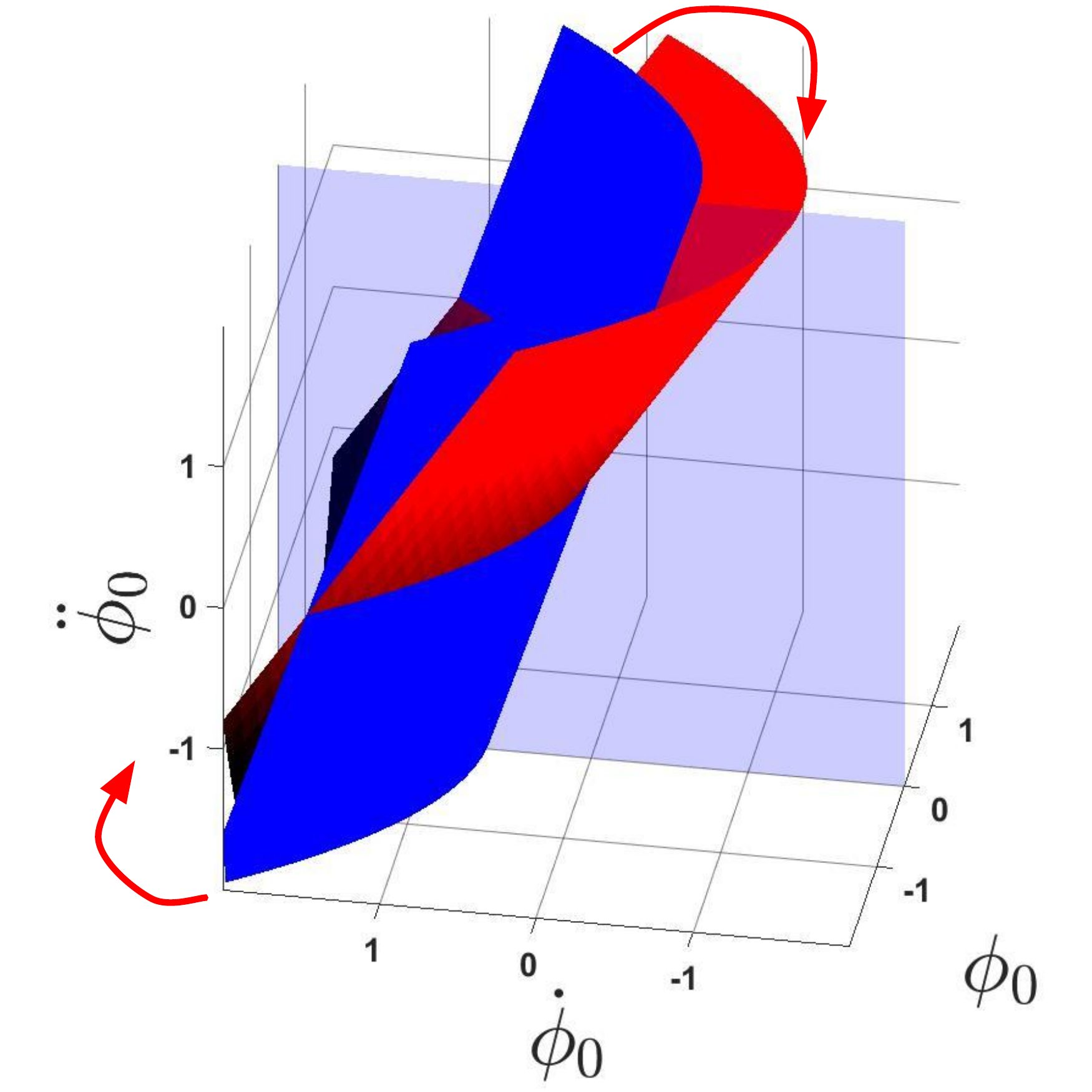}\label{fig:phase_change2}}
\vspace{-10pt}
    \caption{\footnotesize Relationship of $\phi$ when adjusting $\lambda_1$ and $\lambda_2$. \label{fig:phase_change}}
    \vspace{-15pt}
\end{figure}

We observe that by increasing $\lambda_1$, the JSSA active duration time increases. Moreover, the first trigger time is earlier and the last trigger time is delayed.
The minimum relative distance also increases.
These phenomenons indicate that the JSSA is more conservative since it is easier to be triggered and the robot stays further away from the collision.
\Cref{fig:phase_change1} demonstrates the phase portrait change when $\lambda_1$ increases. The transparent blue plane indicates $\phi_0=0$. The blue and the gradient red planes indicate $\phi=0$ with smaller and larger $\lambda_1$ respectively. The red arrows indicate the change of the plane. We can see as $\lambda_1$ increases, the JSSA is easier to be triggered when $\dot{\phi}_0>0$ (larger approaching speed), making the system more conservative. Thus, $\lambda_1$ adjusts the sensitivity to approaching speed.

On the other hand, we observe that the minimum distance decreases as $\lambda_2$ increases from \cref{table:sensitivity}. The average acceleration is suppressed, indicating the system has smoother and less violent behavior.
\Cref{fig:phase_change2} displays the phase portrait when $\lambda_2$ changes. Similarly, the blue transparent plane indicates $\phi_0=0$. The blue and red planes indicate $\phi=0$ with smaller and larger $\lambda_2$ respectively.
We can see as $\lambda_2$ increases, the JSSA is easier to be triggered when $\ddot{\phi}_0\geq 0$ (obstacle accelerating toward the robot), while relaxed when $\ddot{\phi}_0<0$ (obstacle accelerating away from the robot).
Note that in our test case, although the human is approaching the robot, he is decelerating since he stops and turns back (\cref{fig:sim_ca}). Thus, the JSSA is relaxed and a closer distance to the human is allowed.
In addition, the JSSA first trigger time is delayed as $\lambda_2$ increases as shown in \cref{table:sensitivity}.
Thus, $\lambda_2$ adjusts the sensitivity to the relative acceleration.

\begin{figure}
\footnotesize
\centering
\input{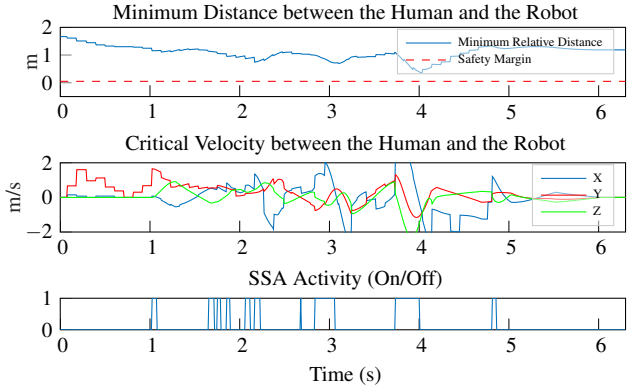}
\vspace{-10pt}
\caption{\footnotesize The simulation profile of the collision avoidance performance of SSA ($\lambda_1=3$) on the FANUC robot.}
    \label{fig:sim_ca_profile_ssa}
\vspace{-15pt}
\end{figure}

\subsection{JSSA vs SSA}
The original SSA \cite{ssa} considers the acceleration-based control system. In this section, we compare the performance of JSSA and SSA, which has $\phi=d_{min}^2-d^2-\lambda_1\dot{d}$, and $\lambda_1=3$.
The SSA is applied on the acceleration to compute the safe acceleration control $\ddot{q}^S$. The jerk control is calculated as $u^S_{k+1}=\frac{\ddot{q}^S_{k+1}-\ddot{q}_k}{\tau}$. 
We saturate the jerk control at the boundary to ensure $u^S\in U$.
The human follows the same trajectory as in \cref{fig:sim_collision_avoidance}.

\Cref{fig:sim_ca_profile_ssa} demonstrates the system profile of the SSA.
By frequently saturating the $u^S$ that violates the jerk bound, the system has a more aggressive behavior ($2^{nd} plot$) and the distance profile fluctuates and has a closer minimum distance (\SI{0.360}{\meter}).
In general, the SSA is less conservative \cite{yao2021safe} comparing to JSSA, which is also shown by having a smaller minimum relative distance.
However, since the system has a decelerating dynamic agent, the SSA (duration: \SI{0.872}{\second}) is easier to be triggered comparing to JSSA (duration: \SI{0.352}{\second}).

\begin{figure}
\footnotesize
\subfigure[]{\includegraphics[width=0.32\linewidth]{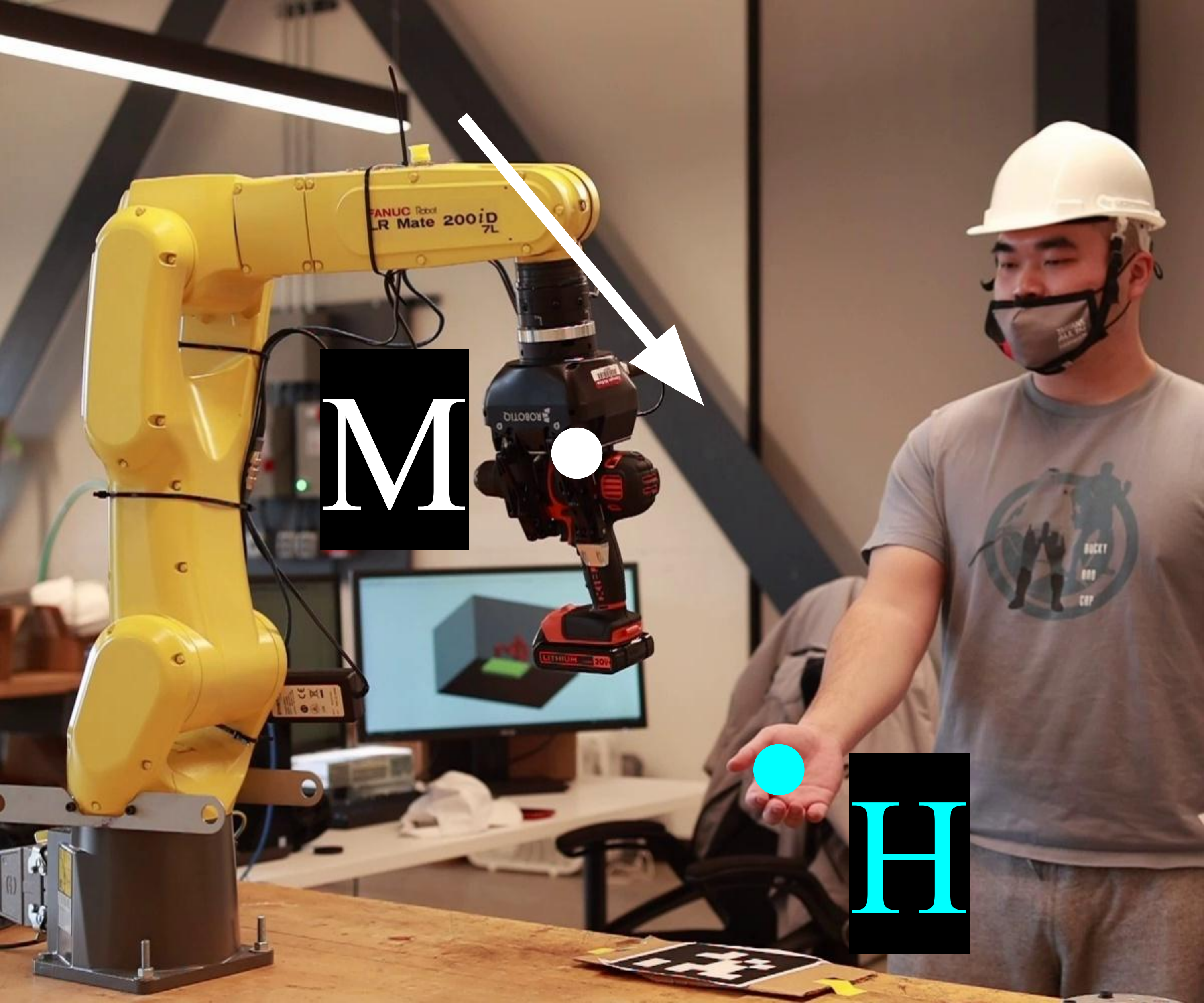}\label{fig:handover1}}\hfill
\subfigure[]{\includegraphics[width=0.32\linewidth]{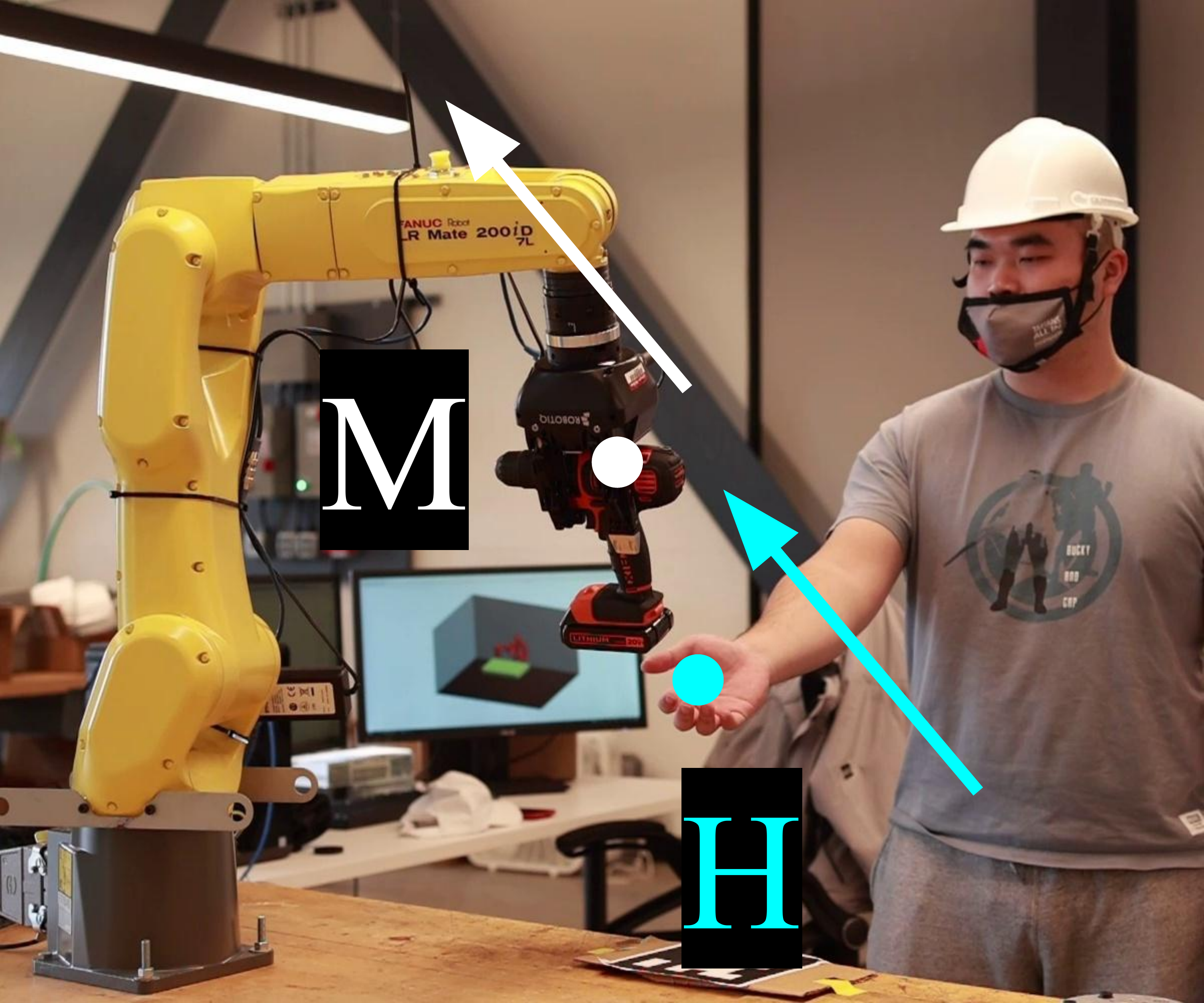}\label{fig:handover2}}\hfill
\subfigure[]{\includegraphics[width=0.32\linewidth]{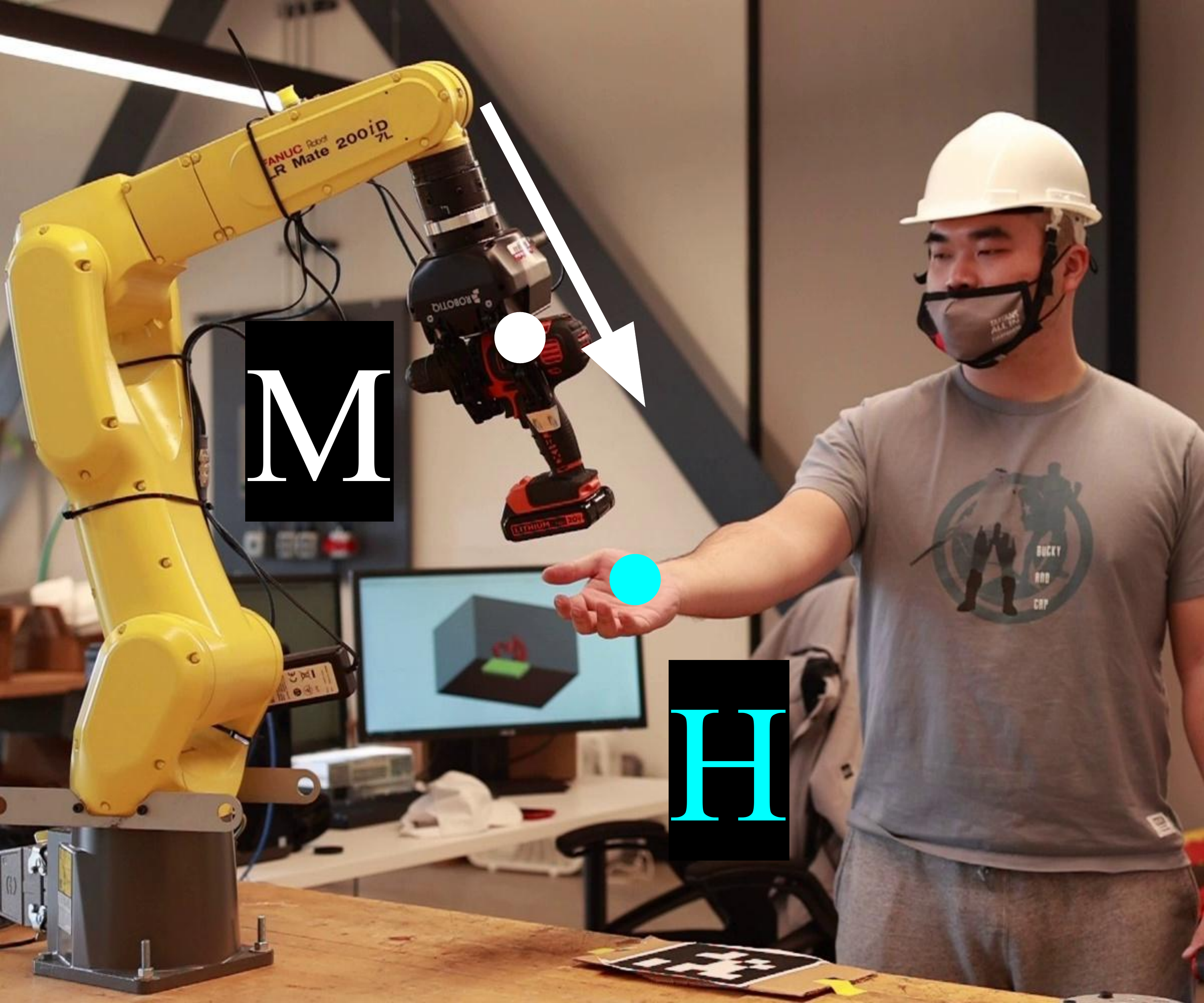}\label{fig:handover3}}
\vspace{-10pt}
    \caption{\footnotesize Visualization of the robot handover. The robot delivers the power drill to the human. $M$ and $H$ indicate the critical point pair. The arrows indicate the motion of the robot and the human. (a) The human places his hand to request the power drill. (b) As the robot delivers, the human raises his hand closer to the robot, and thus, the robot avoids. (c) The robot safely finishes handover with the human's hand higher than the initial position.  \label{fig:real_handover}}
    \vspace{-15pt}
\end{figure}

\begin{figure}
\footnotesize
\centering
\input{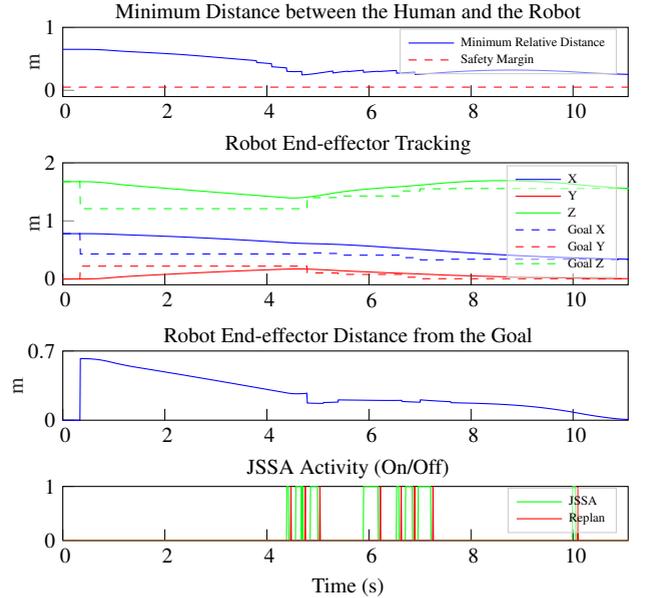}
\vspace{-10pt}
\caption{\footnotesize The system profile of the real-time robot handover performance of JSSA on the FANUC robot.}
    \label{fig:profile_jssa_handover}
\vspace{-15pt}
\end{figure}

\subsection{Real-time Human-Robot Interaction (Handover)}
To further demonstrate the JSSA, we conduct a real-time HRI task: robot handover\footnote[1]{This human-robot interaction study only involves one human performing hand motions. The harm or discomfort
anticipated are no greater than those ordinarily encountered in daily life.}. The robot delivers a power drill to the right hand of the human as shown in \cref{fig:real_handover}, where the right hand is a dynamic obstacle that determines the goal location. The human initially places his hand relatively low (\cref{fig:handover1}). As the robot is delivering, the human moves his hand upward and towards the robot (\cref{fig:handover2}). 
Since the human hand moves, the robot safely avoids the collision and adjusts its motion accordingly to safely finish the handover task (\cref{fig:handover3}).
We track the human pose using a Kinect camera. 

\Cref{fig:profile_jssa_handover} demonstrates the performance of the JSSA in the HRI (robot handover) task.
The first plot indicates that the JSSA maintains a safe distance between the robot and the human even though they approach each other.
From the fourth plot, we observe that the JSSA is triggered at \SI{4.392}{\second}, which is the time that the human starts moving.
The JSSA is continuously triggered as the human continuously moves his hand closer.
We also observe that the replan mechanism is triggered along with the JSSA, which allows the robot to resume the task after JSSA becomes inactive.
From the third plot, we observe that the JSSA makes the robot cautious during delivery as it slightly moves away from the goal from \SI{5}{\second} to \SI{7}{\second}.
When the human hand settles down, the robot then continues to safely finish the task as the end-effector eventually reaches the goal ($2^{nd}$ plot) and the tracking error becomes 0 ($3^{rd}$ plot).

\section{Conclusion} \label{sec:conclusion}

This paper presented a safe set algorithm for jerk-based control systems (JSSA) for interactive industrial robots. We designed JSSA to actively monitor and modify the robot jerk commands to ensure safety in dynamic environments. We synthesized a new safety index that makes a subset of the user-specified safe set forward invariant and ensures control feasibility under robot dynamic constraints. We implemented the JSSA on a FANUC robot where the flexible jerk commands are handled online via a jerk-bounded position controller (JPC). We validated that JSSA was able to actively avoid collision in real-time HRI tasks and performed a sensitivity analysis of system behaviors on hyperparameters. As future work, we aim to extend JSSA to continuous-time systems as well as more complex tasks; derive the formal constraints on the hyperparameters of $\phi$; formally define the forward invariant set given by $\phi$; integrate multiple collision constraints.

\bibliographystyle{plainnat}
% No whitespace allowed in the reference files list, ever.
\bibliography{references}

\end{document}